\documentclass[11pt]{article}

\usepackage[preprint]{acl}

\usepackage{times}
\usepackage{latexsym}

\usepackage[T1]{fontenc}

\usepackage[utf8]{inputenc}

\usepackage{microtype}

\usepackage{inconsolata}

\usepackage{graphicx}
\usepackage{custom}

\title{Taming Object Hallucinations with Verified Atomic Confidence Estimation}

\author{
Jiarui Liu
Weihao Xuan
Zhijing Jin
Mona Diab
}

\author{
 \textbf{Jiarui Liu\textsuperscript{1}},
 \textbf{Weihao Xuan\textsuperscript{2}},
 \textbf{Zhijing Jin\textsuperscript{3}},
 \textbf{Mona Diab\textsuperscript{1}}
\\
 \textsuperscript{1}CMU,
 \textsuperscript{2}The University of Tokyo,
 \textsuperscript{3}The University of Toronto
\\
\href{mailto:jiaruil5@andrew.cmu.edu}{jiaruil5@andrew.cmu.edu}
}

\begin{document}
\maketitle
\begin{abstract}
Multimodal Large Language Models (MLLMs) often suffer from hallucinations, particularly errors in object existence, attributes, or relations, which undermine their reliability. We introduce TACO (Verified Atomic Confidence Estimation), a simple framework that mitigates hallucinations through self-verification and confidence calibration without relying on external vision experts. TACO decomposes responses into atomic queries, paraphrases them to reduce sensitivity to wording, and estimates confidence using self-consistency (black-box) or self-confidence (gray-box) aggregation, before refining answers with a language model. Experiments on five benchmarks (POPE, MME, HallusionBench, AMBER, and MM-Hal Bench) with two MLLMs (\texttt{LLaVA-1.5-7B} and \texttt{CogVLM2}) show that TACO consistently outperforms direct prompting and Visual Contrastive Decoding, reduces systematic biases, and improves confidence calibration, demonstrating its effectiveness in enhancing the faithfulness of MLLMs.
\end{abstract}

\section{Introduction}

\begin{figure*}[t!]
    \centering
    \includegraphics[width=\textwidth]{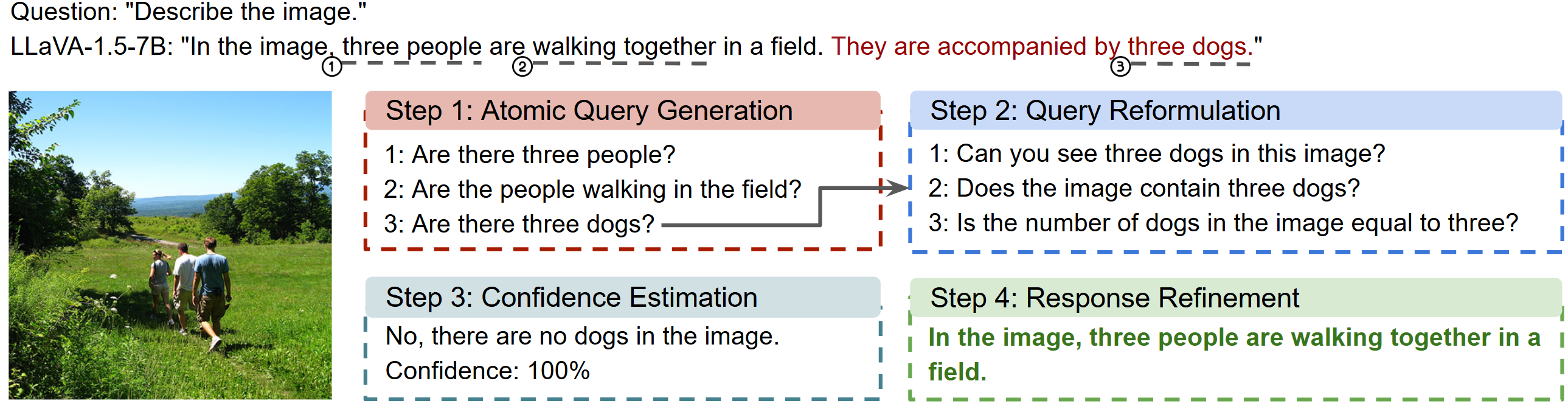}
    \caption{Illustration of the TACO pipeline using a generative example across four steps. 
First, atomic facts are extracted from the query and the original answer, and each fact is framed as a binary atomic query. 
Second, each atomic query is reformulated into multiple semantically equivalent variations to mitigate the over-sensitivity of MLLMs to surface text. 
Third, the MLLM’s responses to these queries are aggregated, and confidence is estimated using either self-consistency (black-box) or self-confidence (gray-box) to select the more reliable answer. 
Finally, an LLM refines the MLLM’s initial response by incorporating the corrected atomic answers.}
    \label{fig:overview}
\end{figure*}

Multimodal Large Language Models (MLLMs) have gained significant attention for their ability to bridge computer vision and natural language processing, excelling in tasks such as Visual Question Answering (VQA) \citep{liu2023visual, yin2024survey, wu2024v}. Despite this progress, MLLMs remain vulnerable to hallucination, producing responses that are unfaithful to the visual input \citep{woodpecker, bai2024hallucination, huang2024opera, favero2024multi}. Unlike hallucinations in purely text-based LLMs, hallucinations in MLLMs often manifest as object hallucinations \citep{pope}, including errors about an object’s existence, attributes, or relations. These errors reduce trustworthiness and limit the adoption of MLLMs in high-stakes applications.

A key cause of hallucination is that MLLMs are overly sensitive to textual variations \citep{chowdhury2025r, ismithdeen2025promptception}. User queries are naturally diverse and fragmentary, yet current models often struggle to align visual content with such variations. Consequently, small differences in wording can lead to inconsistent predictions, undermining reliability \citep{guan2023hallusionbench}. Existing methods to mitigate hallucinations typically rely on external experts such as object detectors or image captioning models \citep{chen24unihd, hallucidoctor}, or on post-hoc calibration strategies \citep{factscore}. However, these approaches introduce dependencies on auxiliary models that may not generalize well and can be computationally expensive.

In this work, we introduce TACO (Verified Atomic Confidence Estimation), a simple yet effective framework to address multimodal hallucinations through self-verification and confidence calibration, without relying on external vision models. TACO operates in four stages: (1) atomic query generation, decomposing user queries and model answers into fine-grained atomic queries that can be verified independently; (2) query reformulation, paraphrasing atomic queries into semantically equivalent variations to mitigate sensitivity to surface-level phrasing; (3) confidence estimation, aggregating responses to reformulated queries using either self-consistency (black-box) or self-confidence (gray-box with logits) estimation to identify the most reliable answer; and (4) response refinement, leveraging a language model to integrate verified atomic answers back into a coherent response. Through this design, TACO systematically detects and corrects hallucinations, making MLLM predictions more consistent and faithful to visual input.

We conduct comprehensive experiments across five benchmarks: POPE \citep{pope}, MME \citep{fu2023mme}, HallusionBench \citep{guan2023hallusionbench}, AMBER \citep{amber}, and MM-Hal Bench \citep{sun2023aligning}, and two state-of-the-art MLLMs, \texttt{LLaVA-1.5-7B} \citep{liu2023visual} and \texttt{CogVLM2} \citep{hong2024cogvlm2}. Results demonstrate that TACO consistently reduces hallucinations in both discriminative and generative tasks, outperforming direct prompting and Visual Contrastive Decoding (VCD). Notably, we find that self-confidence estimation outperforms self-consistency, showing the advantage of gray-box calibration. Beyond benchmark performance, our analysis further reveals how TACO mitigates systematic biases (e.g., “yes”-answer bias) and improves reliability under query reformulations.

In summary, our contributions are threefold:
\begin{enumerate}
    \item We propose TACO, a unified framework for mitigating hallucinations in MLLMs through verified atomic confidence estimation.
    \item We demonstrate that TACO improves calibration in both black-box and gray-box settings, offering insights into the strengths of self-confidence over self-consistency.
    \item We validate TACO across multiple benchmarks and models, showing consistent improvements and providing deeper analysis of its effects on error patterns and biases.
\end{enumerate}

\section{Related Work}

\paragraph{Fact Verification}
Existing fact verification methods for text generation typically follow a multi-stage pipeline that leverages external knowledge bases or domain experts \citep{zhong-etal-2020-reasoning, 10.1162/tacl_a_00454, durmus-etal-2020-feqa, honovich-etal-2022-true}. In VQA, analogous strategies employ external vision experts, such as object detection models \citep{chen24unihd, pope, chair, amber, zhou-etal-2025-m2} or image captioning models \citep{woodpecker, hallucidoctor}, to provide verification evidence. However, these approaches often inherit the limitations of expert outputs, reducing their robustness on out-of-distribution tasks \citep{manakul-etal-2023-selfcheckgpt}. By contrast, self-verification has been explored in the text domain as a means of validating reasoning steps without reliance on external inputs \citep{fabbri-etal-2022-qafacteval, weng-etal-2023-large, miao2023selfcheck, ling2023deductive, zhang2023multicot, li-etal-2024-self}. Inspired by sampling-based self-check techniques in text generation \citep{manakul-etal-2023-selfcheckgpt} and fact extraction in image-to-text tasks \citep{factscore, dsg}, our work investigates the potential of self-verification to mitigate multimodal hallucinations, thereby eliminating the need for external vision experts.

\paragraph{Confidence Calibration}
Modern neural networks are widely recognized for producing poorly calibrated predictions \citep{guo2017calibration, wang-etal-2020-inference, minderer2021revisiting, jiang2021can, xiong2023proximity}. Traditional calibration methods often rely on retraining or the construction of dedicated calibration datasets \citep{NIPS2017_9ef2ed4b, pmlr-v48-gal16, NEURIPS2018_abdeb6f5, yoo2022detection}. With the advent of LLMs, new approaches that avoid full retraining have emerged. These methods can be broadly categorized into verbalization-based techniques \citep{lin2022teaching, zhou-etal-2023-navigating, mielke-etal-2022-reducing, band2024linguistic}, consistency-based techniques \citep{wang2022self, xiong2024can, lyu2024calibrating}, and probability-based techniques \citep{guo2017calibration, zhang2020mix, Malinin2021UncertaintyEI, Kuhn2023SemanticUL, deng2023great, xiong2024can}.

Verbalization-based methods assess whether models can explicitly articulate their confidence, while consistency-based and probability-based methods calibrate output distributions, either without or with access to logits. Typically, multiple generations are sampled using decoding strategies such as temperature scaling \citep{lin2024generating, desai-durrett-2020-calibration}, beam search \citep{Kuhn2023SemanticUL}, or prompt variation \citep{tian-etal-2023-just, xiong2024can, pedapati2024large}. Despite these advances, little work has systematically examined whether model responses can be calibrated through self-estimated confidence derived from question samples in multimodal contexts.

\section{TACO: Verified Atomic Confidence Estimation}

Unlike prior work that primarily validates object-existence hallucinations \citep{amber, pope}, our goal is to detect and correct a broader range of object-level hallucinations in MLLM-generated responses, encompassing inaccuracies in object existence, attributes, and relations \citep{bai2024hallucination}. Our approach consists of four stages: atomic query generation, query reformulation, confidence estimation, and response refinement. Given an image and a query, the MLLM first produces an initial response. We then evaluate its self-consistency or self-confidence on each extracted atomic question, and leverage an LLM to refine the response by resolving any identified hallucinations. An overview of this framework is illustrated in \cref{fig:overview}.

\subsection{Atomic Query Generation}
\label{sec:method_aqg}

\begin{table*}[t!]
    \centering \small
    \begin{tabularx}{\textwidth}{lX}
    \toprule
    \textbf{Hallucination Type} & \textbf{Subcategory} \\
    \midrule
    \textbf{Entity} & Whole (entire entity, e.g., \textit{boy}), Part (part of entity, e.g., \textit{boy's arm}) \\
    \midrule
    \textbf{Attribute} & State (e.g., \textit{happy emoji}), Color (e.g., \textit{white chalk}), Type (e.g., \textit{aviator goggles}), Text rendering (e.g., \textit{text "START"}), Material (e.g., \textit{plastic bowl}), Shape (e.g., \textit{round plate}), Size (e.g., \textit{long bench}), Count (e.g., \textit{three cars}), Texture (e.g., \textit{flattened surface}), Style (e.g., \textit{realistic photo}), Temporal (e.g., \textit{old clock}) \\ 
    \midrule
    \textbf{Relation} & Spatial (e.g., \textit{A behind B}), Action (e.g., \textit{A touches B}) \\
    \bottomrule
    \end{tabularx}
    \caption{Taxonomy used by the atomic query generator to guide the creation of targeted question types. The taxonomy defines core categories of object hallucination, covering entities, attributes, and relations, which can be extended to more comprehensive supersets for broader and benchmark-specific coverage.}
    \label{tab:tuple_taxonomy}
\end{table*}

To comprehensively address different categories of object hallucinations, it is essential to define the types of verification questions that can be generated. Inspired by \cite{dsg}, we introduce a taxonomy of question types in \cref{tab:tuple_taxonomy} and require that all verification questions satisfy two key criteria:

\begin{enumerate}[leftmargin=*, itemsep=0pt, parsep=0pt, topsep=0pt, partopsep=0pt]
    \item Each question must be atomic \citep{dsg}, i.e., it should capture the smallest possible semantic unit. This means focusing on a single atomic fact, as specified by the taxonomy in \cref{tab:tuple_taxonomy}, and ensuring the question is self-contained and answerable without additional context.
    \item Each question must be a positively framed binary question, enabling an unambiguous “yes” or “no” answer.
\end{enumerate}

We employ an LLM to generate atomic questions that provide full semantic coverage of the initial response requiring refinement. To improve this process, we design a two-stage procedure: First, given the user’s question and the MLLM’s initial response, we extract atomic semantic tuples based on the taxonomy. For example, to verify the existence of a truck, the tuple entity–whole (truck) is instantiated from the “Entity–Whole” category. This tuple is then converted into a binary question such as “Is there a truck?” If the original VQA question is already atomic and binary, it is preserved directly as the output. Additional implementation details are provided in \cref{appn:method_aqg_details}.

\subsection{Query Reformulation}
\label{sec:method_aqs}

For each atomic question, we apply question scaling to generate sampled responses for confidence estimation. A straightforward approach would be to use different decoding strategies, such as greedy decoding, beam search, top-$p$ sampling, or temperature scaling. However, these perturbation methods fail to sufficiently explore the MLLM’s output space when applied to binary questions. In contrast, question paraphrasing offers a more effective perturbation strategy, as MLLMs are highly sensitive to syntactic variations in text \citep{ismithdeen2025promptception}. This sensitivity allows paraphrasing to better expose overconfidence and improve calibration. Accordingly, we employ an LLM to paraphrase each atomic question into $n$ variations and evaluate the resulting outputs. Additional implementation details are provided in \cref{appn:method_aqs_details}.

\subsection{Confidence Estimation}
\label{sec:method_ce}

The variance among multiple responses to a given question has been proposed as a proxy for model confidence \citep{xiong2024can}. In this step, given the original image as input, we prompt the MLLM to generate responses to each perturbed atomic question. For each initial atomic question, we collect an answer set $\mathbf{a} = \{ \hat{a}_1, \dots, \hat{a}_n \}$, from which the majority answer is determined as $\bar{a} = \arg\max_{a \in {\text{Yes},\text{No}}} \sum_{i=1}^n \mathbf{1}({\hat{a}_i = a})$ using aggregation functions over candidate answers.

We then propose two methods for estimating and aggregating the MLLM’s self-confidence: \textit{black-box assessment} and \textit{gray-box assessment}. Each method produces a confidence score $\texttt{conf}(q, v, \bar{a})$ for atomic question $q$ given visual input $v$, which is subsequently used to calibrate the reliability of the MLLM’s prediction.

\paragraph{Black-Box Assessment}  
We estimate self-confidence by measuring the self-consistency of candidate answers, without requiring access to the model's internal states or output logits. Following prior work that evaluates agreement between candidate responses $\hat{a}_i$ and the majority answer $\bar{a}$ \citep{wang2022self, lyu2024calibrating}, self-consistency is defined as:
\begin{align}
\label{eq:self_consistency}
    C_{\text{self-consistency}} = \frac{1}{n} \sum_{i=1}^n \mathbb{I}\{\hat{a}_i = \bar{a}\}.
\end{align}

\paragraph{Gray-Box Assessment}  
When output probabilities are available, we can leverage them as uncertainty indicators. For each paraphrased binary question $q_i$ and visual input $v$, we extract the probability $\hat{p}(\hat{a}_i \mid q_i, v)$ assigned to the predicted answer $\hat{a}_i \in \{\text{``Yes''}, \text{``No''}\}$. These probabilities are incorporated as weights in the aggregation function used to assess alignment with the majority answer. By default, we use the mean as the aggregation function, while alternative functions are explored in \cref{sec:as_agg}. The self-confidence score is then computed as:
\begin{align}
\label{eq:confidence}
    C_{\text{self-confidence}} = \frac{1}{n} \sum_{i=1}^n \mathbb{I}\{\hat{a}_i = \bar{a}\} \cdot \hat{p}(\hat{a}_i \mid q_i, v).
\end{align}



\subsection{Response Refinement}
\label{sec:method_rr}

Guided by the atomic verification questions and their answers, we employ an LLM to refine the MLLM’s initial response by incorporating the more confident answer to each atomic query and integrating them into a unified output. Notably, the LLM does not require visual inputs for this step, thereby reducing reliance on the external helper model. If only a single atomic query is generated (i.e., when the original question is already atomic), this refinement step is unnecessary. In our experiments, we use \claudeold{} and \claudenew{} as the LLMs. Additional implementation details are provided in \cref{appn:method_rr_details}.

\section{Experiment Setup}

\begin{table*}[tp]
\centering
\resizebox{\textwidth}{!}{%

\begin{tabular}{ll|ccc|ccc}
\toprule
\multirow{2}{*}{Model} & \multirow{2}{*}{Approach} & \multicolumn{3}{c|}{Accuracy} & \multicolumn{3}{c}{F1} \\
 & & Adversarial & Popular & Random & Adversarial & Popular & Random \\
\midrule
\multirow{6}{*}{\llava{}}  & Direct & 79.60 & 81.73 & 83.47 & 78.36 & 80.17 & 81.71 \\
 & VCD & 81.90 & 84.83 & 86.83 & 81.30 & 83.84 & 85.66 \\
 & \tacos{} w/ \claudeold{} & 83.54 & 86.58 & 88.78 & 83.15 & 85.78 & 87.95 \\
 & \tacof{} w/ \claudeold{} & 84.88 & 87.93 & 89.29 & 84.47 & 87.21 & 88.48 \\
 & \tacos{} w/ \claudenew{} & 84.70 & 87.57 & 89.07 & 84.21 & 86.79 & 88.20 \\
 & \tacof{} w/ \claudenew{} & 84.80 & 87.70 & 89.33 & 84.40 & 86.99 & 88.53 \\
\midrule
\multirow{6}{*}{\cogvlm{}} & Direct & 84.80 & 85.90 & 88.53 & 84.17 & 85.06 & 87.55 \\
 & VCD & 85.87 & 87.20 & 89.97 & 85.11 & 86.37 & 89.04 \\
 & \tacos{} w/ \claudeold{} & 85.32 & 85.90 & 86.84 & 83.61 & 84.10 & 84.99 \\
 & \tacof{} w/ \claudeold{} & 85.64 & 86.53 & 87.42 & 84.00 & 84.84 & 85.70 \\
 & \tacos{} w/ \claudenew{} & 84.70 & 85.50 & 86.13 & 82.53 & 83.40 & 83.98 \\
 & \tacof{} w/ \claudenew{} & 84.77 & 85.67 & 86.43 & 82.67 & 83.64 & 84.39 \\

\bottomrule
\end{tabular}

}
\caption{Results on POPE. \textit{Direct} denotes the direct sampling baseline, and \textit{VCD} refers to the visual contrastive decoding baseline \citep{vcd}.}

\label{tab:pope}
\end{table*}

\begin{figure*}[h!]
    \centering
    \begin{minipage}{0.48\textwidth}
        \includegraphics[width=\textwidth]{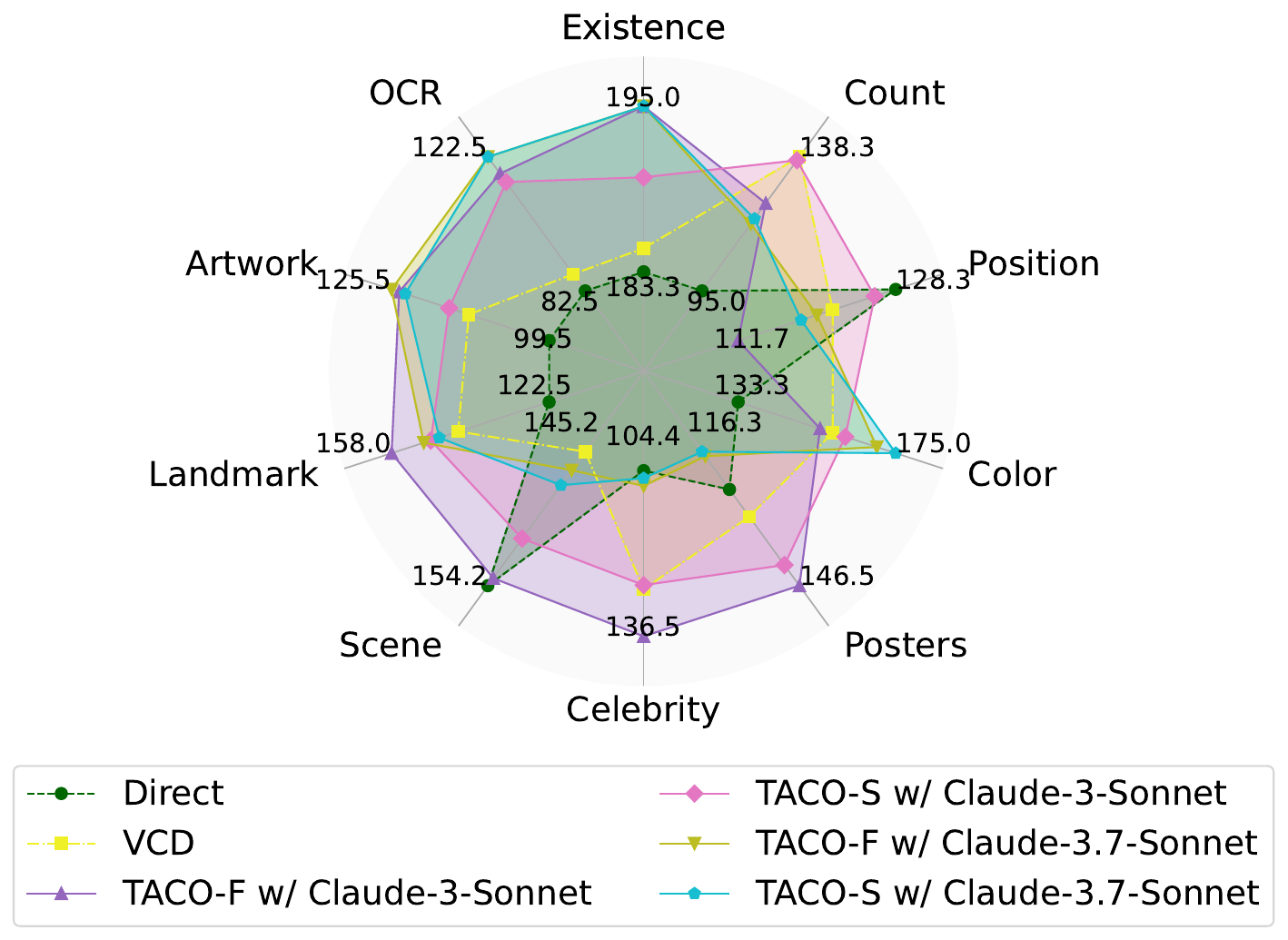}
        \subcaption{\texttt{LLaVA-1.5-7B} on MME.}
        \label{fig:mme_llava15}
    \end{minipage}
    \hfill
    \begin{minipage}{0.48\textwidth}
        \includegraphics[width=\textwidth]{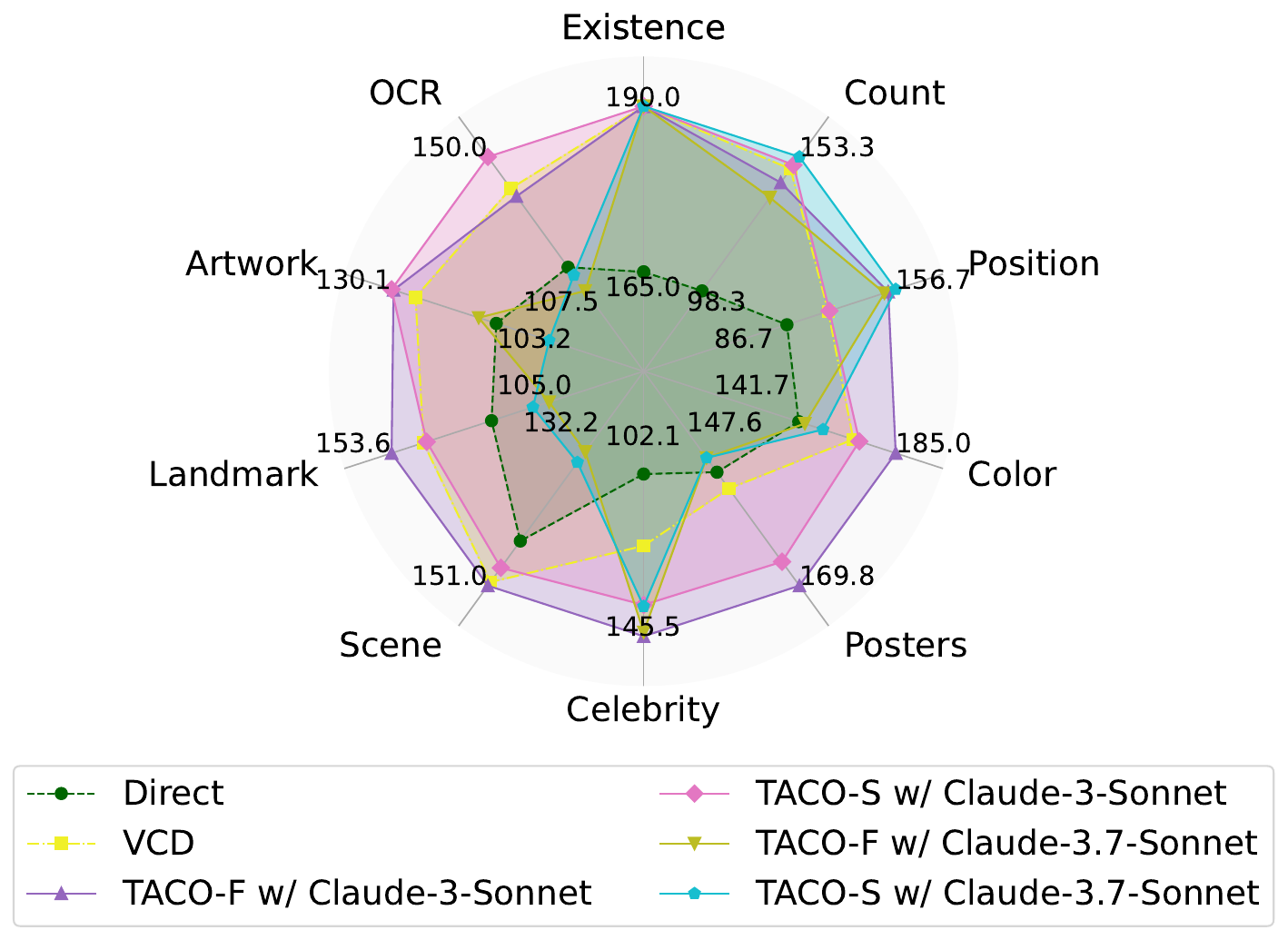}
        \subcaption{\texttt{CogVLM2} on MME.}
        \label{fig:mme_cogvlm2}
    \end{minipage}
    \caption{Results on the MME benchmark for both models. Subfigures show perception-task performance of (a) \texttt{LLaVA-1.5-7B} and (b) \texttt{CogVLM2}.}
    \label{fig:mme}
\end{figure*}

We evaluate our confidence calibration approach on two state-of-the-art MLLMs, \llava{} and \cogvlm{}. Experiments are conducted across three discriminative benchmarks: POPE \citep{pope}, MME \citep{fu2023mme}, and HallusionBench \citep{guan2023hallusionbench} as well as two generative benchmarks: AMBER \citep{amber} and MM-Hal \citep{sun2023aligning}. We compare TACO against the following baselines: (1) direct prompting of the MLLM, and (2) VCD, which applies contrastive decoding by comparing the model’s responses with those generated from a perturbed image input \citep{vcd}. We refer to our self-consistency approach as \tacos{} and our self-confidence approach as \tacof{}.
Further implementation details are provided in \cref{appn:experiment_details}.

\paragraph{POPE}
POPE \citep{pope} introduces a polling-based query method to evaluate an MLLM’s ability to answer object-existence questions in images. The task is framed as binary classification using yes/no questions, which improves both stability and flexibility. The benchmark is constructed from 500 MSCOCO images with an equal number of positive and negative ground-truth examples, yielding 9,000 questions. These are evenly distributed across three settings: \textit{random}, \textit{popular}, and \textit{adversarial}, for generating negative examples. We report accuracy and F1 score.

\paragraph{MME}
The MME benchmark evaluates the perception and cognition capabilities of MLLMs across 14 tasks with 2374 examples \citep{fu2023mme}. Consistent with our definition in \cref{sec:method_aqg}, each question is binary, focusing on an atomic fact derived from an image. Following \cite{woodpecker, vcd}, we restrict our evaluation to perception tasks in order to specifically examine object hallucination. We report the score following \citet{fu2023mme} based on accuracy.

\paragraph{HallusionBench}
HallusionBench is a benchmark designed to evaluate the failure modes of MLLMs \citep{guan2023hallusionbench}. It contains 1,129 questions divided into two categories: \textit{Visual Dependent} and \textit{Visual Supplement}. \textit{Visual Dependent} questions require visual information for accurate answers, while \textit{Visual Supplement} questions can be answered without it, with the visual input serving only as additional context or correction. This design enables assessment of both visual reasoning ability and the interplay between language priors and visual content. All questions are binary under our definition. We report question-pair accuracy (qAcc), figure accuracy (fAcc), and overall accuracy (aAcc) as evaluation metrics.

\paragraph{AMBER} 
AMBER is an LLM-free, multi-dimensional benchmark designed to evaluate hallucinations in MLLMs across both generative and discriminative tasks \citep{amber}. In our work, we use its 1,004 generative questions to assess the effectiveness of our atomic question generation approach. Evaluation is based on four metrics: CHAIR, Cover, Hal, and Cog. CHAIR measures the frequency of hallucinated objects in responses; Cover measures object coverage; Hal measures the proportion of responses containing hallucinations; and Cog measures whether MLLM hallucinations resemble patterns observed in human cognition.

\paragraph{MM-Hal}
MM-Hal is a benchmark specifically designed to evaluate hallucinations in MLLMs \citep{sun2023aligning}. Unlike prior benchmarks that focus on general response quality or restrict evaluation to yes/no questions, MM-Hal employs open-ended, realistic questions to better capture hallucination phenomena in practical settings. It consists of 96 image–question pairs across 8 categories and 12 object topics, targeting common failure modes such as incorrect attributes, adversarial objects, counting, comparison, spatial relations, environment, and holistic descriptions. Evaluation is performed with GPT-4o as the judge, comparing model responses against human-written references to determine whether hallucinations are present.

\section{Results}

\paragraph{POPE}
The experimental results on POPE across the random, popular, and adversarial settings are shown in \cref{tab:pope}. Our approach consistently outperforms both baselines in terms of accuracy and F1 score across all subtasks on \texttt{LLaVA-1.5-7B}, while achieving comparable performance on \texttt{CogVLM2}. Importantly, self-confidence estimation consistently surpasses self-consistency estimation across all subtasks and models, demonstrating that leveraging the model’s probabilistic distribution leads to more effective confidence calibration. The performances of \claudeold{} and \claudenew{} are largely comparable.

\paragraph{MME} \cref{fig:mme} shows radar plots of evaluation results across MME perception task subsets. Unlike benchmarks focused solely on object existence, MME also evaluates attribute-level hallucinations, providing a broader assessment of model reliability. Our methods consistently reduce hallucinations across different subsets and MLLMs. Consistent with the POPE results, self-confidence estimation generally outperforms self-consistency in terms of overall accuracy. By contrast, the VCD approach yields only marginal gains.



\paragraph{HallusionBench} Our evaluation shows that both of our approaches outperform baseline methods across all accuracy metrics. The self-consistency and self-confidence estimation methods achieve largely comparable results, while VCD does not yield clear improvements on this benchmark. This is because HallusionBench provides a more rigorous and adversarial evaluation of VQA dependencies on visual content. In such cases, VCD often fails when the perturbed image does not offer sufficiently informative contrastive cues to guide the model’s attention.

\begin{table*}[tp]
\centering

\resizebox{\linewidth}{!}{%
\begin{tabular}{ll|ccccc}
\toprule
Model & Approach & qAcc ($\uparrow$) & fAcc ($\uparrow$) & Easy aAcc ($\uparrow$) & Hard aAcc ($\uparrow$) & aAcc ($\uparrow$) \\
\midrule
\multirow{6}{*}{\llava{}} & Direct & 17.58 & 19.36 & 42.42 & 43.72 & 49.60  \\
& VCD & 16.92 & 18.79 & 41.10 & 42.33 & 49.25 \\
 & TACO-S w/ Claude-3-Sonnet & 19.56 & 21.39 & 46.37 & 46.28 & 50.49  \\
 & TACO-F w/ Claude-3-Sonnet & 14.51 & 22.54 & 45.93 & 48.60 & 51.20  \\
 & TACO-S w/ Claude-3.7-Sonnet & 20.88 & 25.14 & 49.67 & 48.60 & 55.09  \\
 & TACO-F w/ Claude-3.7-Sonnet & 20.44 & 26.30 & 48.35 & 50.00 & 55.62  \\
 \midrule
\multirow{6}{*}{\cogvlm{}} & Direct & 21.98 & 21.68 & 50.55 & 42.33 & 53.06  \\
& VCD & 21.32 & 23.12 & 52.75 & 41.40 & 52.88  \\
 & TACO-S w/ Claude-3-Sonnet & 30.11 & 32.08 & 53.85 & 56.28 & 60.50  \\
 & TACO-F w/ Claude-3-Sonnet & 30.55 & 36.13 & 56.48 & 61.86 & 63.86  \\
 & TACO-S w/ Claude-3.7-Sonnet & 24.84 & 27.46 & 50.33 & 55.35 & 59.26  \\
 & TACO-F w/ Claude-3.7-Sonnet & 24.62 & 25.43 & 50.99 & 54.19 & 59.26  \\
\bottomrule
\end{tabular}
}
\caption{Results on HallusionBench.}
\label{tab:hallusionbench}
\end{table*}

\paragraph{AMBER}
The results for both models in \cref{tab:amber} show that TACO outperforms the direct prompting baseline on the CHAIR, Hal, and Cog metrics, while maintaining comparable object coverage as measured by COVER. Notably, the hallucination rate decreases substantially, with minimal loss of useful information.

\begin{table}[tp]
\centering
\resizebox{\linewidth}{!}{%
\begin{tabular}{llcccc}
\toprule
Model & Approach & CHAIR ($\downarrow$) & COVER ($\uparrow$) & Hal ($\downarrow$) & Cog ($\downarrow$) \\
\midrule
\multirow{3}{*}{\llava{}} & Direct & 11.7 & 51.1 & 49.5 & 4.4 \\
 & TACO-S w/ \claudenew{} & 6.5 & 48.5 & 29.1 & 1.9 \\
 & TACO-F w/ \claudenew{} & 6.4 & 49.0 & 28.8 & 2.0 \\
\midrule
\multirow{3}{*}{\cogvlm{}} & Direct & 11.3 & 61.9 & 50.9 & 4.1 \\
 & TACO-S w/ \claudenew{} & 7.6 & 59.1 & 36.6 & 2.1 \\
 & TACO-F w/ \claudenew{} & 7.7 & 59.0 & 37.1 & 2.2 \\
\bottomrule
\end{tabular}
}
\caption{Results on AMBER. For all metrics except \textit{COVER}, lower values indicate better performance.}
\label{tab:amber}
\end{table}

\begin{table}[tp]
\centering
\resizebox{\linewidth}{!}{%
\begin{tabular}{llcc}
\toprule
Model & Approach & Average Score ($\uparrow$) & Hallucination Rate ($\downarrow$) \\
\midrule
\multirow{3}{*}{\llava{}} & Direct & 1.85 & 0.69 \\
 & TACO-S w/ \claudenew{} & 2.25 & 0.58 \\
 & TACO-F w/ \claudenew{} & 2.53 & 0.52 \\
 \midrule
\multirow{3}{*}{\cogvlm{}} & Direct & 2.71 & 0.53 \\
 & TACO-S w/ \claudenew{} & 2.59 & 0.50 \\
 & TACO-F w/ \claudenew{} & 2.59 & 0.51 \\

\bottomrule
\end{tabular}
}
\caption{Results on MM-Hal Bench. Lower values indicate better performance for hallucination rate.}
\label{tab:mmhal}
\end{table}

\paragraph{MM-Hal} On MM-Hal in \cref{tab:mmhal}, we find that TACO improves performance for \llava{} but not for \cogvlm{}. Notably, \cogvlm{} outperforms \llava{} by a significant margin. We attribute this to the strong baseline performance of CogVLM2 on this task, where the involvement of an auxiliary language model may introduce negative effects when the original MLLM already possesses sufficient capability to answer the questions. 

\section{Discussion}

Based on our experiments across both discriminative and generative benchmarks, we highlight two key findings: (1) TACO consistently outperforms both direct sampling and visual contrastive decoding baselines, demonstrating its effectiveness as a confidence calibration method. (2) Self-confidence estimation surpasses self-consistency estimation, showing that gray-box assessments provide deeper insight into an MLLM’s calibrated confidence.

Below, we further analyze statistics from POPE and HallusionBench to shed light on the underlying behavior of our method.

\paragraph{Does the model exhibit greater consistency across reformulated queries when producing correct answers?}

To examine whether the dispersion of model outputs (i.e., disagreement among paraphrased responses) is linked to prediction reliability, we analyzed the per-question variance of binary answers on POPE. For each question, we computed the proportion of ``yes'' responses and derived the variance $p(1-p)$ as a measure of uncertainty. The majority vote was then compared against the ground truth to assess correctness.

We tested whether variance differed significantly between correctly and incorrectly answered questions using Welch’s two-sample $t$-test, and verified robustness with the non-parametric Mann–Whitney $U$ test. We also quantified the relationship between variance and correctness using the point-biserial correlation. Results show that incorrect predictions exhibit substantially higher mean variance than correct ones, with both statistical tests confirming significance and a negative correlation observed between variance and correctness.

These findings support the hypothesis that greater disagreement among paraphrased answers is predictive of reduced majority-vote accuracy. They further suggest that MLLMs are more sensitive to syntactic variation when less confident, consistent with our intuition.

\begin{table}[tp]
\centering
\resizebox{\linewidth}{!}{%
\begin{tabular}{l|ccc|ccc}
\toprule
\multirow{2}{*}{Metric} & \multicolumn{3}{c|}{\llava{}} & \multicolumn{3}{c}{\cogvlm{}} \\
& Random & Popular & Adv. & Random & Popular & Adv. \\
\midrule
T-Test p-value & 3.1e-25 & 3.3e-21 & 3.0e-23 & 2.0e-21 & 6.8e-22 & 3.5e-22 \\
Mann-Whitney U p-value & 4.4e-88 & 2.1e-58 & 4.5e-53 & 3.3e-63 & 7.3e-61 & 3.1e-56 \\
\bottomrule
\end{tabular}
}
\caption{Statistical significance tests comparing the variance $p(1-p)$ of atomic answer distributions between correctly and incorrectly answered questions on POPE. Reported values are $p$-values from Welch’s two-sample $t$-test and the Mann–Whitney $U$ test across random, popular, and adversarial settings for \llava{} and \cogvlm{}. Lower $p$-values indicate stronger evidence that variances differ between correct and incorrect answers.}
\label{tab:stats_test_var}
\end{table}

\paragraph{What bias does TACO correct?}

Prior work has shown that MLLMs often exhibit a bias toward answering “Yes,” particularly in existence and attribute verification tasks \citep{guan2023hallusionbench}. This tendency stems from poor confidence calibration and leads models to over-predict object presence or attributes even when they are absent.

To evaluate whether TACO mitigates this bias, we analyzed the ratio of “Yes” responses on HallusionBench, comparing our approach with baseline methods. Specifically, we measured the deviation in \textit{Yes} prediction percentages from the reference values reported by \citet{guan2023hallusionbench}.

Our results show that TACO effectively reduces the systematic “Yes” bias observed in the baselines, producing more balanced predictions between “Yes” and “No.” In particular, our methods substantially lower the internal bias of both models, as reflected in improvements on the Pct Diff and FP Ratio metrics. These findings indicate that TACO not only calibrates confidence more accurately but also helps correct the inherent response imbalance in MLLMs.

\begin{table}[tp]
\centering
\resizebox{\linewidth}{!}{%
\begin{tabular}{l|cc|cc}
\toprule
\multirow{2}{*}{Approach} & \multicolumn{2}{c|}{\llava{}} & \multicolumn{2}{c}{\cogvlm{}} \\
 & Pct Diff (\textasciitilde 0) & FP Ratio (\textasciitilde 0.5) & Pct Diff (\textasciitilde 0) & FP Ratio (\textasciitilde 0.5) \\
\midrule
Direct & 0.12 & 0.62 & 0.17 & 0.68 \\
VCD & 0.14 & 0.63 & 0.19 & 0.70 \\
TACO-S & 0.00 & 0.50 & -0.07 & 0.41 \\
TACO-F & 0.02 & 0.52 & -0.08 & 0.40 \\
\bottomrule
\end{tabular}
}
\caption{Analysis of Yes/No prediction bias on HallusionBench. 
We report the \textit{Yes Percentage Difference} (Pct Diff; closer to 0 indicates balanced Yes/No predictions) 
and the \textit{False Positive Ratio} (FP Ratio; closer to 0.5 indicates reduced bias toward predicting “Yes”). 
Results show that both TACO-S and TACO-F substantially reduce the Yes-bias compared to the direct and VCD baselines for both \llava{} and \cogvlm{}.}
\label{tab:yes_bias}
\end{table}

\paragraph{Which aggregation function is better for self-confidence estimation: MEAN or MAX?}
\label{sec:as_agg}

\begin{figure}[ht]
    \centering
    \includegraphics[width=\linewidth]{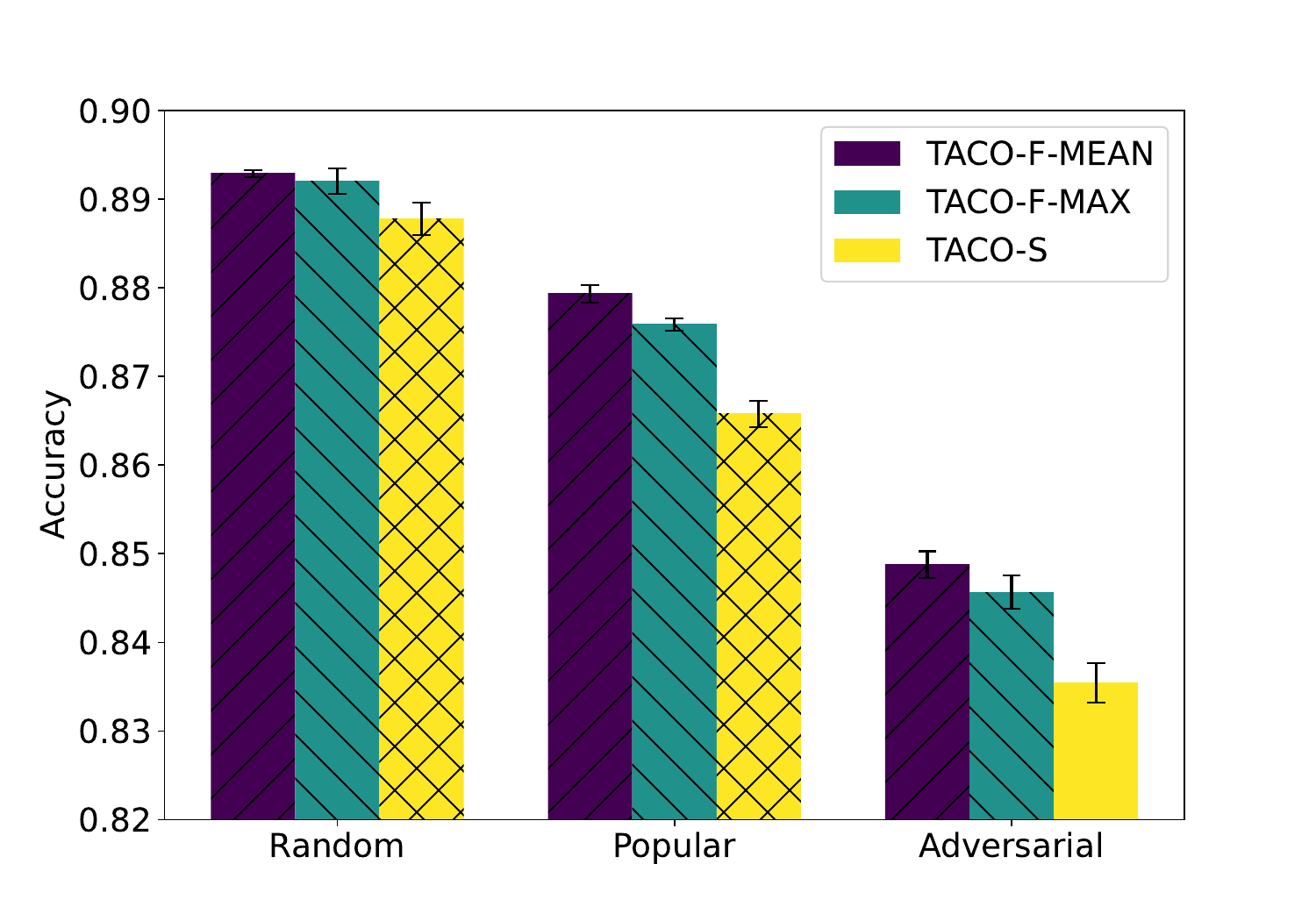}
    \caption{Comparison of aggregation functions for self-confidence estimation on POPE using \llava{}.}
    \label{fig:agg_func_llava}
\end{figure}

We compare two aggregation strategies for self-confidence estimation: MAX and MEAN. The MAX function selects the larger probability between predicting Yes’’ and No’’ across all paraphrased atomic questions, while the MEAN function computes the average probability of each answer over all paraphrases. As shown in \cref{fig:agg_func_llava}, MEAN consistently provides more reliable confidence estimates than MAX, leading to superior calibration performance.



\paragraph{Can LLMs handle negative questions?}

During our experiments, we observed an interesting phenomenon: current MLLMs struggle with negative queries. Both \llava{} and \cogvlm{} frequently misinterpret questions such as “Isn’t there a ball in the image?” or “Are there no balls in the image?”, often producing inconsistent or random answers, even when they correctly answer the corresponding positive query. This suggests that their basic linguistic reasoning capabilities remain limited, highlighting the need for improved training strategies to strengthen textual understanding and better align language with visual inputs.

\section{Conclusion}

In this work, we presented a sampling-based confidence calibration framework to address object hallucinations in VQA tasks. Our four-step pipeline systematically generates atomic questions, reformulates them into multiple variations, estimates the MLLM’s confidence, and refines its initial response through calibrated feedback. Experiments on multiple benchmarks and state-of-the-art MLLMs show that question paraphrasing is an effective strategy for sampling diverse generations, while LLM-assisted atomic fact extraction and question formulation substantially reduce hallucinations in generative settings. These results highlight the potential of verified atomic confidence estimation as a lightweight yet powerful approach for improving the faithfulness and reliability of MLLMs.

\section*{Limitations}

While our study demonstrates the effectiveness of TACO in mitigating multimodal hallucinations, several limitations remain that open up directions for future work.

First, our framework is evaluated primarily on a selection of widely used benchmarks. Although these cover both discriminative and generative tasks, they may not fully capture the diversity of real-world multimodal scenarios, such as open-domain dialogue, video understanding, or interactive systems. Extending the evaluation to broader and more dynamic datasets would help verify the generalizability of our approach.

Second, our method currently focuses on hallucinations at the object level: existence, attributes, and relations. While these represent the most common and impactful error modes, hallucinations can also manifest in more abstract forms, such as commonsense reasoning or causal inference. Incorporating atomic query generation for higher-level semantic reasoning remains an interesting direction for future research.

Finally, while we deliberately avoid reliance on external vision experts to keep the framework lightweight and generalizable, there may still be scenarios where integrating complementary signals from specialized vision modules could further enhance robustness. Exploring hybrid approaches that combine self-verification with external guidance in a controllable way may provide a balance between autonomy and reliability.

\section*{Ethical Considerations}

Our work addresses the problem of hallucinations in MLLMs, with the goal of improving the faithfulness and reliability of model outputs. While this research aims to reduce the risks associated with inaccurate or misleading responses, mitigating hallucinations does not guarantee the elimination of all errors. Models calibrated with TACO may still produce inaccurate or biased outputs, particularly when operating on out-of-distribution data. Users should be aware that even with improved confidence estimation, MLLMs should not be blindly trusted in safety-critical domains such as healthcare, law, or autonomous systems without human oversight.

Our framework relies on large-scale pretrained MLLMs and LLMs, which themselves may encode societal biases present in their training data. Although TACO helps calibrate confidence and correct factual inconsistencies, it does not directly address biases or harmful stereotypes inherent to the underlying models.

Additionally, improving faithfulness may inadvertently increase user trust in MLLMs. While this is a desirable outcome for reliability, it also carries the risk of over-reliance, particularly if users assume that calibrated models are universally correct. It is important that system designers clearly communicate residual limitations and provide mechanisms for human-in-the-loop verification.

\bibliography{custom}

\appendix

\section{Appendix}

\subsection{Atomic Query Generation Details}
\label{appn:method_aqg_details}

Atomic query generation consists of two steps. First, we prompt an LLM to generate tuples following the provided taxonomy. The prompt is shown below. We provide five-shot examples.

\begin{lstlisting}
Task: Based on the example input questions, the example output tuples, and the provided tuple taxonomy below, generate skill-specific tuples to help verify and refine the answer of the last input question. 

Requirements:
1. Ensure the generated tuples fully capture the factual information of the input question, with each tuple representing a distinct atomic and positive statement. Subjective elements in the initial answer should be disregarded.
2. If the input question is irrelevant to any category, output "None."
3. You must remove any negative words including "not" and "no" from your generation regardless of whether it will result in the opposite meaning.
4. Do not generate trivial tuples about the image itself such as "entity - whole (image)".
5. Each tuple should be output in the following format: id | tuple

Tuple taxonomy:
```
Entity relationships:
* entity - whole
* entity - part

Attribute relationships:
* attribute - state
* attribute - color
* attribute - type
* attribute - text rendering
* attribute - material
* attribute - shape
* attribute - size
* attribute - count
* attribute - texture
* attribute - style
* attribute - temporal

Relations:
* relation - spatial
* relation - action

Miscellaneous:
* other - other
```
\end{lstlisting}

Second, we prompt the same LLM to convert these tuples into atomic queries. The corresponding prompt is shown below. We conduct experiments with two LLMs, \claudeold{} and \claudenew{}. For both models, the temperature is set to 0 and the maximum output length to 1000 tokens. This decoding configuration is applied consistently across all uses of the LLM. We provide two shot examples.

\begin{lstlisting}
Task: Given the example input questions, skill-specific tuples, and the example output of generated binary questions, re-write each tuple from the last example into a standalone, positively framed natural language binary question. 

Requirements:
1. Each binary question should be non-trivial for a vision model to verify. Exclude trivial tuples that do not help in verifying and refining the initial answer.
2. Each binary question should be self-contained and answerable independently, without requiring knowledge of other binary questions.
2. Generate one binary question only for the two or more tuples sharing the same meaning or the opposite meaning.
3. Ensure the generated questions fully capture the factual information of the input question. Create additional binary questions if they are helpful and complementary for refining the initial answer.
4. Treat conditional statements or given information in "Question:" as context that you don't need to ask questions from.
5. You must generate positively framed questions and remove any negative words including "not" and "no" from your generation regardless of whether it will result in the opposite meaning. For example, instead of generating "is this artwork not created by Jacob?", you should always ask its corresponding positive question "is this artwork created by Jacob?"
output format: id | question
\end{lstlisting}

\subsection{Query Reformulation Details}
\label{appn:method_aqs_details}

For each atomic query, we generate nine reformulated variations using the same LLM as above. The corresponding prompt is shown below. We provide two-shot examples.

\begin{lstlisting}
Paraphrase the following question about an image maintaining the exact same meaning. You must keep the entity names in the paraphrased questions the same as in the input question to prevent any ambiguity. Ensure each generated question is easily understandable and can be answered with "yes" or "no." Generate 10 distinct paraphrased versions of the question.
    
Input question:
```
{question}
```

Directly provide your paraphrased questions in a numbered list without any explanations.
\end{lstlisting}

\subsection{Response Refinement Details}
\label{appn:method_rr_details}

The following prompt is used to guide the LLM in refining the MLLM’s answer.

\begin{lstlisting}
Given a VQA question-answer pair, refine the model's initial answer using the context of verification questions and their ground truth answers. Preserve the model's answer if the verification context confirms that the final answer is correct, even if the model's reasoning is flawed. Only revise the model's answer if the verification context provides highly specific and directly relevant evidence that the final answer itself is incorrect. If no sufficiently relevant verification questions are available, return the initial answer as the output. Ensure that all output text is derived from the initial answer or the provided context; do not generate any new, unverified information.

Question: "{question}"

Model's initial answer: "{answer}"

Verification context:
```
{verification_qa}
```

Provide only the revised answer without any explanation or additional text.
\end{lstlisting}

\subsection{Experiment Details}
\label{appn:experiment_details}

We conduct our experiments on 8 NVIDIA A100 80G GPUs, totaling approximately 1,000 GPU hours. For all experiments involving MLLMs, results are averaged over three random seeds, and we report the mean performance. We set the decoding temperature to 0.6 to reduce repetitions in text generation and limit the maximum number of new tokens to 1,024. For VCD, we use $\alpha = 1$, $\beta = 0.1$, and 500 noise steps. The total API cost for LLM usage was approximately \$1000.

We use the official implementations of \llava{} \footnote{\url{https://github.com/haotian-liu/LLaVA}} and \cogvlm{} \footnote{\url{https://github.com/zai-org/CogVLM2}}.


\subsection{Dataset details}

All datasets used in this work are subject to their respective licenses and are employed strictly for research purposes, consistent with their original intended use. The datasets contain only VQA examples and do not include any personally identifiable information or offensive content.

\subsection{Usage of AI Assistants}

We use AI assistants solely to correct grammar and improve clarity of writing.

\end{document}